\documentclass[12pt]{article}
\usepackage{lipsum}
\usepackage{arxiv}

\usepackage[utf8]{inputenc} 
\usepackage[T1]{fontenc}    
\usepackage{hyperref}       
\usepackage{url}            
\usepackage{booktabs}       
\usepackage{amsfonts}       
\usepackage{amsmath}        
\usepackage{nicefrac}       
\usepackage{microtype}      
\usepackage{graphicx}
\usepackage{amsbsy}
\usepackage{mathrsfs}
\usepackage{xcolor}
\usepackage{multirow}
\usepackage{array}
\usepackage{tablefootnote}
\DeclareMathAlphabet{\mathscrbf}{OMS}{mdugm}{b}{n}
\graphicspath{ {./figures/} }
\providecommand{\keywords}[1]
{
  \small	
  \textbf{\textit{Keywords---}} #1
}

\title{Real-time estimation of overt attention from dynamic features of the face using deep-learning }

\author{
Aimar Silvan Ortubay\\
  Biomedical Engineering\\
  City College of New York\\
  \texttt{asilvanortubay@ccny.cuny.edu} \\
   \And
Lucas C. Parra \\
  Biomedical Engineering\\
  City College of New York\\
  \texttt{parra@ccny.cuny.edu} \\
   \And
Jens Madsen\\
    Biomedical Engineering\\
   City College of New York\\  \texttt{jmadsen@ccny.cuny.edu} \\
}

\begin{document}
\maketitle
\begin{abstract}
Students often drift in and out of focus during class. Effective teachers recognize this and re-engage them when necessary. With the shift to remote learning, teachers have lost the visual feedback needed to adapt to varying student engagement. We propose using readily available front-facing video to infer attention levels based on movements of the eyes, head, and face. We train a deep learning model to predict a measure of attention based on overt eye movements. Specifically, we measure Inter-Subject Correlation of eye movements in ten-second intervals while students watch the same educational videos. In 3 different experiments (N=83) we show that the trained model predicts this objective attention metric on unseen data with $R$²=0.38, and on unseen subjects with $R$²=0.26-0.30. The deep network relies mostly on a student's eye movements, but to some extent also on movements of the brows, cheeks, and head.  In contrast to Inter-Subject Correlation of the eyes, the model can estimate attentional engagement from individual students' movements without needing reference data from an attentive group. This enables a much broader set of online applications. The solution is lightweight and can operate on the client side, which mitigates some of the privacy concerns associated with online attention monitoring. GitHub implementation is available at \url{https://github.com/asortubay/timeISC}.
\end{abstract}

\section{Introduction}

The COVID-19 pandemic forced a rapid shift to online learning, shaking up traditional in-person education. Students suddenly found themselves in a less engaging environment while they struggled to maintain focus \cite{fmg15,scyzy24,r17}. This shift induced by the pandemic accelerated an ongoing trend to use pre-recorded video as instructional material. In such a context, an important challenge is to keep learners engaged despite the missing or limited interactions between students and teachers \cite{wslfm14,h08}. This highlights a growing need for tools that sense and adapt to student engagement in remote settings. 

Currently, the most common attention-monitoring tool is observational checklists, where students or tutors assess engagement through questionnaires, self-reports, or rating scales \cite{dlpg14,ot10}. However, these methods are intrusive, time-consuming, subjective, and fail to capture the natural fluctuations in engagement over time \cite{ot10,dml19}.  Other solutions look at the total number of clicks \cite{kgsmgm14} and playtime \cite{gupta2022daiseeuserengagementrecognition} but are not real-time \cite{dml19}, and it is less certain that they capture attention to the video content. Automatic tools and datasets have emerged to address these limitations, often using facial expressions and postures to infer attentional states \cite{gupta2022daiseeuserengagementrecognition,kaur2018predictionlocalizationstudentengagement}. However, these tools rely on large manually annotated datasets, which are expensive to create and prone to subjective bias \cite{bammp19}.

In search of more objective metrics, some studies have explored using neural activity \cite{co18,cp16,hnlfm04}, physiological signals like heart rate \cite{po21}, along behavioral signals like eye-tracking \cite{mp22,mjgsp21}, to estimate cognitive states of arousal and attention. They have shown that attentive students' neural, physiological, or behavioral responses synchronize with other people when they engage with the same stimulus. Inter-Subject Correlation (ISC) is a robust and scalable metric to measure attentional state and predict student performance. However, acquiring these signals generally requires expensive equipment and controlled settings \cite{dml19,rbtk16}. Additionally, measuring ISC requires reference data from an attentive group watching identical video material, which limits the widespread adoption of such techniques.

In this study, we focus on detecting student engagement by analyzing the dynamics of facial and head movements captured through a webcam, without needing any special setup or calibration. To alleviate the need for a reference signal, we train a deep learning model to predict  ISC of the eye movements as an objective measure of attention while students watch instructional videos. We demonstrate that model predictions generalize to new participants and different videos, providing a real-time estimate of engagement. Our method provides a practical, scalable, and anonymous way to monitor student engagement continuously, while also addressing some of the challenges found in previous research. We also analyzed which specific face or head movements contribute to the engagement estimate.

\section{Methods}

One of the main challenges in measuring student engagement is to cope with the complex, natural environment of remote education. Our approach aims to mimic this setting closely as students watch a stimulus on a screen—such as a lecture by a professor or a short educational video. To measure their attention non-intrusively, we use MediaPipe -- a free, easily accessible software API -- that tracks facial movements without requiring any calibration.

\subsection{Datasets}

Our method was developed using existing webcam recordings from the experiment described by Madsen et al. \cite{mjgsp21}. This study investigated whether the ISC of eye gaze and pupil size, measured by an eye-tracking device, could predict test performance and distinguish between attentive and distracted states. In the original study, participants watched instructional videos while their eye movements were tracked. Our study leveraged the webcam recordings that captured participants' faces and torsos from a camera positioned above the presentation screen, simulating an online learning environment. The recordings were made at 60 fps and aligned with the stimulus presentation using timestamps. To validate our attention estimate across different scenarios, we used data from three different experiments. In Experiment 1, we tested our model by predicting attention from participants who were part of the training cohort, but on a third of the watching time that was left out of the training set. Experiment 2 assessed the generalization to a new cohort of subjects who watched the same videos as those in training. In Experiment 3, we evaluated the model’s performance on an entirely different cohort of subjects watching different videos.


\begin{table}[h]
\caption{Summary of experimental data available}
\centering
\begin{tabular}{|c|c|c|c|c|c|}
\hline
\multirow{2}{*}{\textbf{Dataset}} & \multicolumn{2}{c|}{\textbf{Subjects}} & \multicolumn{2}{c|}{\textbf{Stimuli}} & \textbf{Webcam data} \\ \cline{2-6} 
                                  & \textit{N} & \textit{(M/F)} & \textit{N} & \textit{Type} & \textit{Duration (hours)} \\ \hline
Experiment 1                      & 26         & (10/16)      & 5          & Same          & 9.18                     \\ \hline
Experiment 2                      & 29         & (10/19)      & 5          & Same          & 11.14                    \\ \hline
Experiment 3                      & 28         & (8/20)       & 6          & Different     & 15.45                    \\ \hline
\end{tabular}
\label{table1}
\end{table}

In Experiments 1 and 2, participants watched the same five short educational videos (length 2.4-6.5 min, M=4.1 min, SD=2.0 min). In Experiment 3, a third cohort watched six different videos (length 4.2-6 min, M=5.15 min, SD=57 s). All stimuli are publicly available on YouTube. Together, these experiments provided a total of 35.77 hours of recordings across eleven different educational videos. For more details on the stimuli, experimental conditions, and subject information, readers are referred to the original study \cite{mjgsp21}.

\subsection{Face tracking and landmark normalization}

We used the MediaPipe FaceLandmarker model \cite{m24,grishchenko2020attentionmeshhighfidelityface}, a computer vision model, often referred to as FaceMesh, that estimates face landmark locations and geometry. Designed for mobile devices, this lightweight model requires only a webcam frame as input and provides real-time performance in face landmark estimation. While originally developed for augmented reality applications, its precision makes it ideal for analyzing facial movements and dynamics \cite{ssn21}. The model outputs 478 3D landmarks (in relative webcam frame coordinates) and 52 blendshapes, representing the intensities (ranging from 0 to 1) of specific facial movements. The MediaPipe model was run on the raw webcam videos using Python 3.11 and MediaPipe v0.10.14. 

Since the landmark locations are extracted in video-frame coordinates, converting them to a common 3D space is necessary to enable between-participant comparison. This is done by computing an affine transformation from the screen coordinates ($X$) to the Canonical Face Model ($C$) \cite{m24,grishchenko2020attentionmeshhighfidelityface}, which represents a standard 3D face in metric space. The affine transformation matrix ($R$) is calculated as the linear least-squares solution that maps the extracted landmarks to the canonical face:
\begin{equation}
 R = (X^T \cdot X)^{-1} \cdot X^T \cdot C 
\end{equation}
Then, the landmarks are transformed to the new common metric space (Y) using the following equation:
\begin{equation}
 Y= X \cdot R
\end{equation}

\begin{figure}[ht]
\begin{center}
\includegraphics[width=0.5\columnwidth]{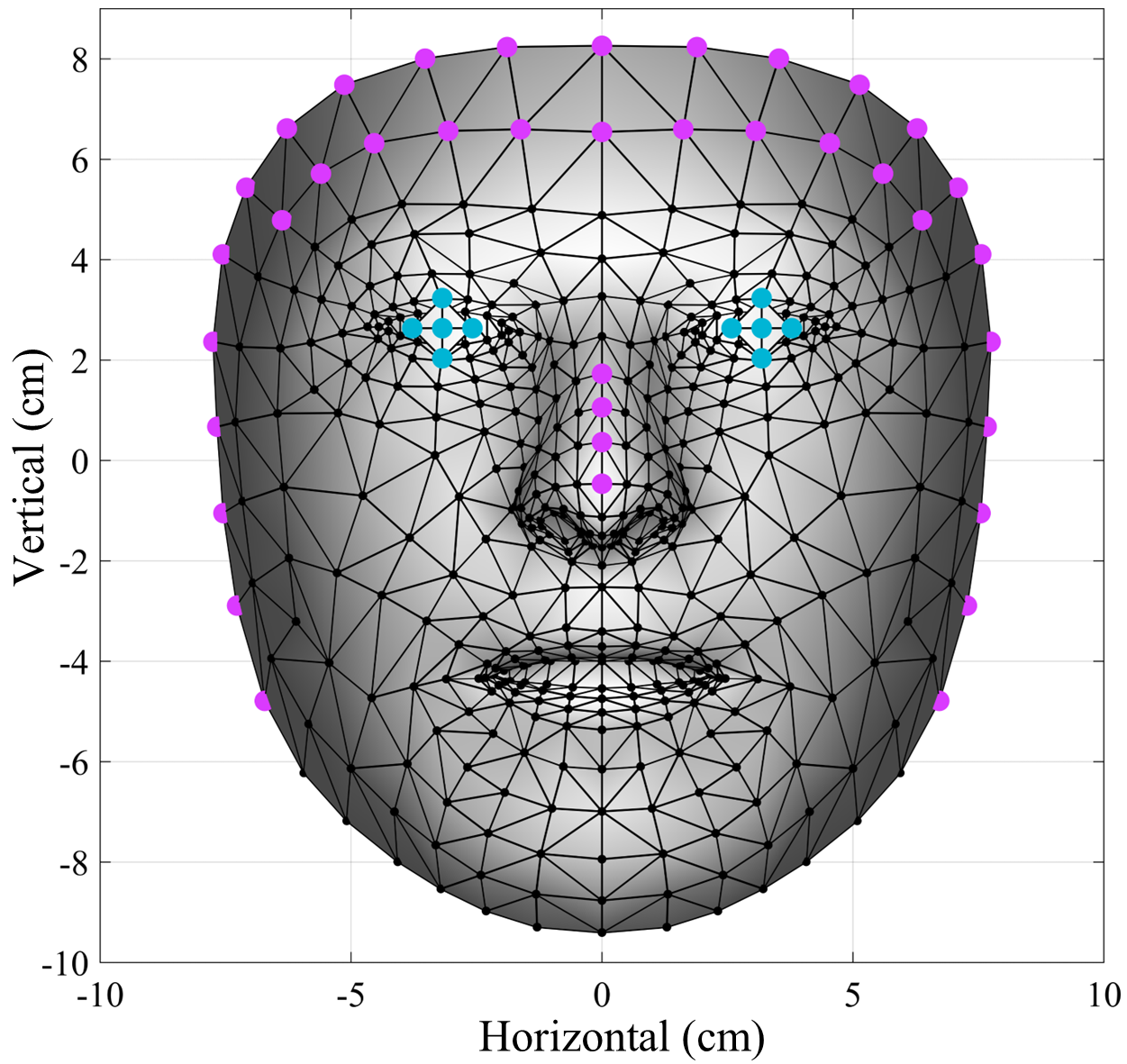}
\end{center}
\definecolor{Mycolor1}{HTML}{FF00FF}
\definecolor{Mycolor2}{HTML}{00FFFF}
\caption{Canonical face showing the vertices and 478 landmarks tracked by Mediapipe FaceMesh, colored points depict the landmarks used in affine transformation (\textcolor{Mycolor1}{magenta}) and the tracked iris (\textcolor{Mycolor2}{cyan})}
\label{FaceMesh}
\end{figure}

Matrices $X$, $C$, and $Y$ are arranged as samples by dimensions. To ensure that relevant movement information is preserved in the process, the landmarks used to solve the least-squares equation are selected based on their expected stability during natural face movements. These stable landmarks, depicted in Figure~\ref{FaceMesh}, help maintain the integrity of the transformation. Additionally, the affine matrix $R$ contains valuable information about the participants’ heads’ rotation and position.

\subsection{Time-resolved Inter-Subject Correlation}

\begin{figure}[ht]
\begin{center}
\includegraphics[width=1\columnwidth]{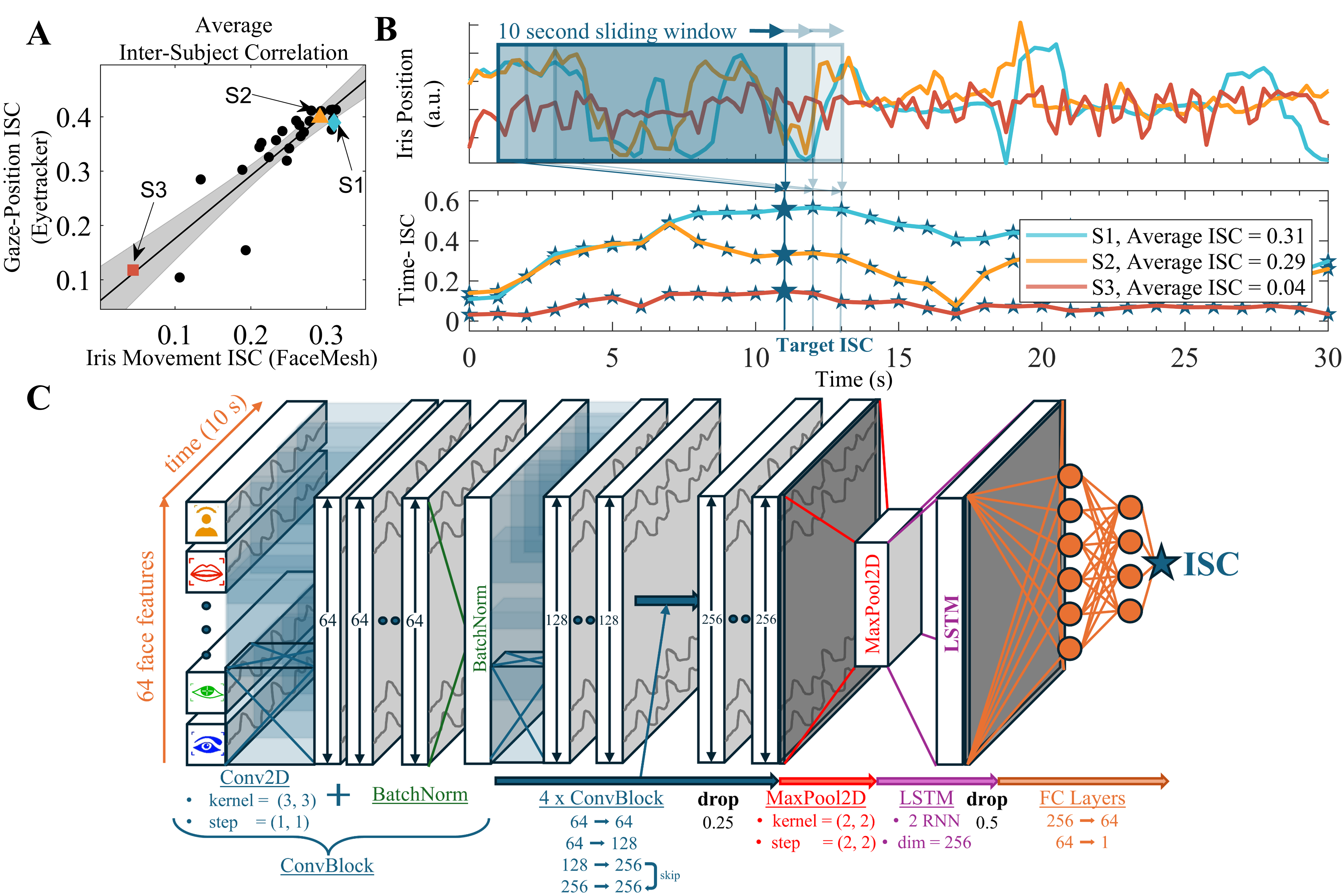}
\end{center}
\caption{Engagement prediction framework: 10 seconds of dynamic face features predict a single Inter-Subject Correlation (ISC) value. A: Iris movements tracked by FaceMesh exhibit ISC similar to that of gaze position captured by a research-grade eyetracker ($r(24)=0.89, p=1.2e^{-9}$), data from Exp. 1 (N=26). B: 10-second windows of FaceMesh iris movements are used to compute time-resolved ISC, in one-second steps. Two attentive subjects (S1 and S2) show similar iris movements and have higher instantaneous and average ISC, compared to a non-attentive subject (S3). C: An example of a single instance prediction, the model will learn to predict an ISC value from 10 seconds of preceding dynamic face and head movements (64 features) of a single student. The best-performing model has a series of spatiotemporal convolutions, ReLU activations, batch normalization, and max-pooling layer, followed by an LSTM and fully connected layers.  
}
\label{framework}
\end{figure}

Previous experiments using research-grade eye-tracking devices \cite{mjgsp21} demonstrated that eye-gaze synchrony and pupil size were strongly linked to the participants’ attentional states, with substantially higher ISC during an attentive state, as compared to a distracted state. We found that iris movements tracked by FaceMesh exhibited ISC values similar to those of gaze position captured by the eyetracker, without the need for calibration (Figure~\ref{framework}A). However, continuously tracking attention with this approach required pre-recorded data from multiple participants watching the identical video. To address this, we developed a framework that predicts a student’s ISC at any given moment using only 10-second segments of the face and head movement data, effectively removing the need for other participants' data. Previous analysis indicated that relevant modulation of ISC with attention happened for slow eye movements (< 2 Hz, peaking at 0.1 Hz) \cite{mjgsp21}. Based on this, we generated training and testing datasets by computing time-resolved ISC through a sliding 10-second window with one-second steps, sampling the FaceMesh iris landmark  data at 4 Hz (Figure~\ref{framework}A). This sampling rate was selected to capture relevant frequencies while keeping computational costs low.


For each window, we calculated the pairwise Pearson correlation between all pairs of participants. The ISC value for each participant was then obtained by averaging the pairwise correlation values between that participant and all others in the same set, excluding self-correlation. Negative correlations were set to zero, and the average ISC values across iris landmarks were used to represent overall iris synchrony. This process was repeated for each window, yielding a dynamic, one-second resolution signal of ISC over time, with each time-ISC value used as the regression target.

We defined ten seconds of 64 face features as predictors (Figure~\ref{framework}C), including the 52 FaceMesh-extracted blendshapes tracking facial movements and the affine matrix $R$, which tracks head movements. Ten seconds had been previously reported as the appropriate length for engagement assessment by humans \cite{wslfm14}. We use the same preceding ten seconds’ worth of face tracking data as those used to compute ISC from the iris movements.

\subsection{Deep-learning estimation of time-resolved ISC} 

We adapted several deep learning architectures for our regression task, such as MLPs, LSTMs, 1D-Convolutions, and a hybrid model combining 2D-Convolutions with an LSTM. We also adopted various state-of-the-art models previously used for multivariate time-series classification \cite{ffwim19}. The models were implemented using Pytorch 2.1.2, trained with mean-squared error as the loss function, and optimized using the Adam optimizer for 150 epochs. The best-performing model was selected based on its validation loss (Table~\ref{table2}).

For training, validation, and testing the different experiments outlined in Table~\ref{table1} were treated as separate datasets, to test the generalizability of the results to different scenarios. We used the data from twenty participants from Experiment 1 for training, three for validation and selection of the models, and three for testing on unseen subjects (Table~\ref{table2}). To externally validate the test results from Experiment 1 on a broader set of subjects watching the same stimulus videos, we tested the model on subjects from Experiment 2 (N=29). To test if the model was capturing individual-specific clues from seen participants, we tested the predictions on the left-out data from the training participants (N=20). Finally, we assessed the overall generalizability of the model to unseen subjects and stimuli by testing on Experiment 3 (N=28). The different dataset test prediction errors were computed using Mean Absolute Error (MAE), which we compared to a baseline naive model that simply predicted the average ISC value of the training set. Additionally, the coefficient of determination ($R$²) was used to evaluate how effectively the models captured the inherent variance in the time-resolved correlations. Due to the variability between participants, both MAE and $R$² were computed individually for each participant and then averaged to report overall performance (Table~\ref{table3}). This also allowed for paired t-test performance comparisons between the model predictions and the baseline.

To better understand the features driving the attention-level predictions, we conducted a feature suppression study by zeroing out specific features from the set of 64 predictors. We then measured the resulting increase in prediction error, providing insights into the relative importance of different facial and head movements for assessing ISC (Figure~\ref{results}). This offered a degree of explainability regarding the features most relevant to the model's predictions.

\section{Results \& Discussion}

We found that iris movements, tracked solely via webcam, captured gaze synchrony, comparable to research-grade eye-tracking devices (Figure~\ref{framework}A). This allows us to distinguish between highly engaged students and those less attentive. This was observed in both instantaneous and average correlation levels (Figure~\ref{framework}B). This proves that our approach offers a calibration-free alternative to traditional eye-tracking methods, potentially providing a cost-effective solution for remotely predicting student performance \cite{mjgsp21}.


\begin{table}[h]
\caption{Model selection based on Experiment 1}
\centering
\begin{tabular}{|c|c|c|c|}
\hline
\multirow{2}{*}{\textbf{Model}} & \multicolumn{3}{c|}{\textbf{\textit{R}²}} \\ \cline{2-4} 
                                  & \textit{Train ($N=20$)} & \textit{Val ($N=3$)} & \textit{Test ($N=3$)} \\ \hline
    MLP & 0.649 & 0.3154 & 0.182 \\ \hline
    Simple LSTM & 0.5077 & 0.3795 & 0.3139 \\ \hline
    1D CNN & 0.6182 & 0.4012 & 0.3097 \\ \hline
    2D CNN + LSTM & 0.9719 & \textbf{0.4118} & 0.3621 \\ \hline
    1D ResNet & 0.9632 & 0.3065 & 0.1987 \\ \hline
    Encoder\cite{spk18} & 0.961 & 0.3659 & 0.3423 \\ \hline
    Time Le-Net \cite{gmt} & 0.3155 & 0.33 & 0.1895 \\ \hline
    MCDCNN \cite{zlcgz16} & 0.4299 & 0.353 & 0.3227 \\ \hline
\end{tabular}
\label{table2}
\end{table}

The best-performing model in predicting ISC from individual face data (Table~\ref{table2}) combined spatiotemporal convolutions with a recurrent layer (Figure~\ref{framework}C), highlighting the importance of temporal dynamics in capturing subtle facial movements \cite{asc05}. As this is the first attempt to predict synchrony levels from a single subject, there are no direct comparisons to previous studies. However, across all scenarios tested, including predictions on unseen subjects and new video stimuli, the model prediction error was consistently lower than simply predicting the average ISC value (Table~\ref{table3}). This indicates that the model effectively captures relevant facial dynamics linked to attention, suggesting its broader applicability to different individuals and visual materials. The explained variance ($R$²) decreased for unseen subjects (Table~\ref{table3}), reflecting how both humans and artificial intelligence models struggle with interpreting unfamiliar faces \cite{lb20,cgyg14}.


\begin{table}[h]
\caption{Model selection based on Experiment 1}
\centering
\begin{tabular}{|c|c|c|c|c|c|}
\hline
\multirow{2}{*}{\textbf{Model}} & \multirow{2}{*}{\textbf{N}} & \multicolumn{3}{c|}{\textbf{Mean Absolute Error}} & \multirow{2}{*}{\textbf{\textit{R}²}} \\ \cline{3-5}
                        & & \textit{Baseline} & \textit{Model} & \textit{difference} &\\ \hline
                                  
    Exp. 1 (same subj., different stims) & 20 & 0.157 ± 0.052 & 0.110 ± 0.034 & -29.96\%* & 0.384 \\ \hline
    Exp. 2 (same subj., different stims) & 29 & 0.149 ± 0.011 & 0.126 ± 0.012 & -14.95\%* & 0.259** \\ \hline
    Exp. 3 (different subj., different stims) & 28 & 0.153 ± 0.012 & 0.124 ± 0.015 & -19.03\%* & 0.301** \\ \hline
\multicolumn{4}{l}{\small *Significant decrease in error on model vs. baseline ($p$<0.0001, paired $t$-test).} \\
\multicolumn{4}{l}{\small **$R$² is significantly lower than in Exp. 1 ($p$<0.05, paired $t$-test).} \\
\end{tabular}
\label{table3}
\end{table}

\begin{figure}[ht]
\begin{center}
\includegraphics[width=0.65\columnwidth]{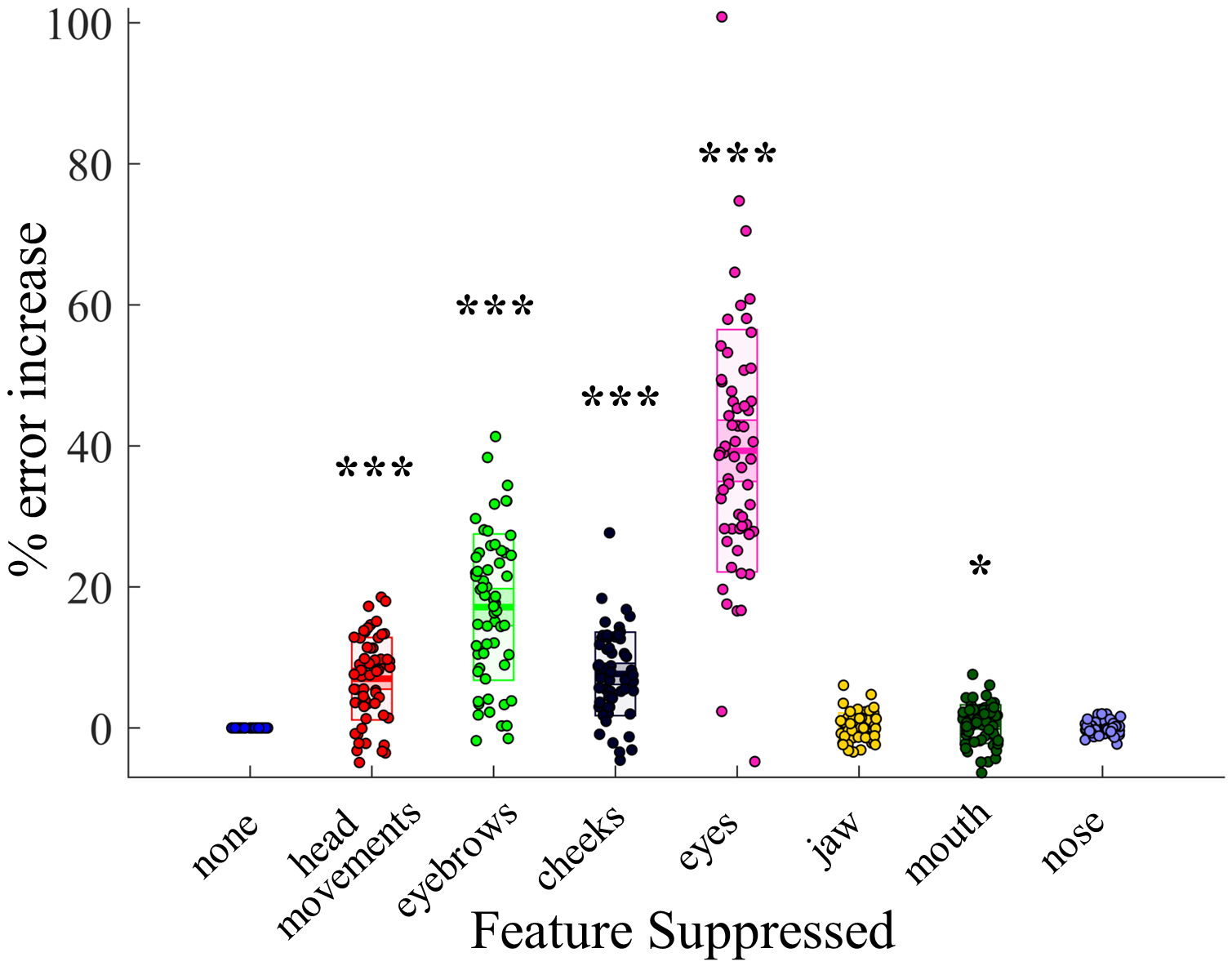}
\end{center}
\definecolor{Mycolor1}{HTML}{FF00FF}
\definecolor{Mycolor2}{HTML}{00FFFF}
\caption{The attention prediction relies heavily on eye movements, and to a lesser extent on the head, eyebrows, cheeks, and mouth movements.  \%-change in Mean Absolute Error per participant on feature suppression study, model errors significantly increase (***$p<0.001$, *$p<0.05$, paired-tailed $t$-test), for Exp. 2 \& 3 participants (df=N=57) after zeroing out specific predictor features vs. unchanged (‘none’)}
\label{results}
\end{figure}

Finally, the feature suppression analysis revealed that the model relies primarily on head and eye movements, with other facial regions also contributing (Figure~\ref{results}). The prediction error significantly increased when specific head or facial movement information was removed, aligning with findings that both human observers and automated systems use eye movements and head pose to assess engagement levels \cite{wslfm14}.

In all, our model offered a promising framework for continuous assessment of overt attention. While we extracted features from pre-recorded webcam videos, in a remote education setting the MediaPipe framework could be run on the client side via a web interface with low computational requirements, with real-time performance \cite{g24}. This would provide a secure setting where only the engagement estimate would be transmitted, for consenting participants.

\section{Conclusion}

Students sometimes struggle to maintain focus, especially when faced with unengaging online lectures. This study analyzes face and head movements to estimate students’ attentional states. Unlike in prior attention-prediction frameworks, we introduced an objective attention metric using Inter-Subject Correlation (ISC), giving us insights into how students synchronize and follow along with the educational material. Using readily available face-tracking tools, we developed a model that predicts engagement in real-time, offering continuous attention monitoring during remote lectures. Our model’s performance generalized beyond the original stimuli, demonstrating its applicability to diverse educational videos and audiences. However, results indicate that the model partially relies on individual-specific facial cues, which could affect its generalizability across unseen subjects. With a larger training dataset, we anticipate improvements in both the model's accuracy and broader applicability, ultimately providing an effective, real-time, objective metric for assessing student engagement and potentially predicting their performance.
\newpage
\bibliographystyle{plos2015} 
\bibliography{references}

\end{document}